\documentclass[10pt,a4paper]{scrartcl}
\pdfoutput=1
\usepackage[utf8]{inputenc}
\usepackage[english]{babel}
\usepackage{cmbright}
\usepackage{url}
\usepackage[]{geometry}
\usepackage[T1, OT1]{fontenc}

\usepackage{amsmath}
\usepackage{enumitem}
\usepackage{float}
\usepackage{natbib}

\usepackage{xkeyval}
\usepackage{xstring}
\usepackage{iftex}
\usepackage{microtype}
\usepackage{etoolbox}
\usepackage{booktabs}
\usepackage{refcount}
\usepackage{totpages}
\usepackage{environ}
\usepackage{setspace}
\usepackage{textcase}
\usepackage[bookmarksnumbered,unicode,hidelinks]{hyperref}
\usepackage{graphicx}
\usepackage[prologue]{xcolor}
\usepackage{manyfoot}
\usepackage{cmap}
\usepackage{caption, float}
\usepackage{array}
\usepackage{authblk}
\usepackage{framed}

\title{\vspace{-1cm}The Proof is in the Almond Cookies}
\subtitle{A Case Study on Narrative-Based Understanding of Recipes}
\author[1]{Remi van Trijp}
\author[2]{Katrien Beuls}
\author[3]{Paul Van Eecke}
\affil[1]{\textit{Sony Computer Science Laboratories Paris,  France}}
\affil[2]{\textit{Facult\'{e} d'informatique, Universit\'{e} de Namur, Belgium}}
\affil[3]{\textit{Artificial Intelligence Laboratory, Vrije Universiteit Brussel, Belgium}}
\date{\vspace{-1.5cm}}

\def\keywordname{{\bfseries \emph Keywords:}}%
\def\keywords#1{\par\addvspace\medskipamount{\rightskip=0pt plus1cm
\def\and{\ifhmode\unskip\nobreak\fi\ $\cdot$
}\noindent\keywordname\enspace\ignorespaces#1\par}}

\begin{document}

\maketitle

\begin{abstract}
\noindent This paper presents a case study on how to process cooking recipes (and more generally, how-to instructions) in a way that makes it possible for a robot or artificial cooking assistant to support human chefs in the kitchen. Such AI assistants would be of great benefit to society, as they can help to sustain the autonomy of aging adults or people with a physical impairment, or they may reduce the stress in a professional kitchen. We propose a novel approach to computational recipe understanding that mimics the human sense-making process, which is narrative-based. Using an English recipe for almond crescent cookies as illustration, we show how recipes can be modelled as rich narrative structures by integrating various knowledge sources such as language processing, ontologies, and mental simulation. We show how such narrative structures can be used for (a) dealing with the challenges of recipe language, such as zero anaphora, (b) optimizing a robot's planning process, (c) measuring how well an AI system understands its current tasks, and (d) allowing recipe annotations to become language-independent.
\end{abstract}

\keywords{narrative, grounded language understanding, human-centric AI, mental simulation}

\vspace{.5cm}

\noindent
\begin{framed}
\noindent
Please cite as:\\van Trijp,  R., Beuls, K. \& Van Eecke, P. 2024. The Proof is in the Almond Cookies.  In Steels, L. \& Porzel, R.  (eds).  \textit{Narrative-based Understanding of Everyday Activities: A Cookbook.} Venice: Venice International University.  Pages 59--77.
\end{framed}

\section{Introduction}

This paper explores what kind of grounded language processing model is needed for enabling robots or computational cooking assistants to support human chefs in the kitchen. Such human-centric AI assistants would be of great benefit for society because they could sustain the autonomy of aging adults or people with a physical impairment, or they could reduce the pressure on professional chefs who have to work in high-stress situations. We propose a novel approach to computational recipe understanding that mimics the narrative-based sense-making process of humans \citep{bruner1991narrative}, which may lead to more intuitive and meaningful human-robot interactions.

\begin{figure}
\centering
  \includegraphics[width=\textwidth]{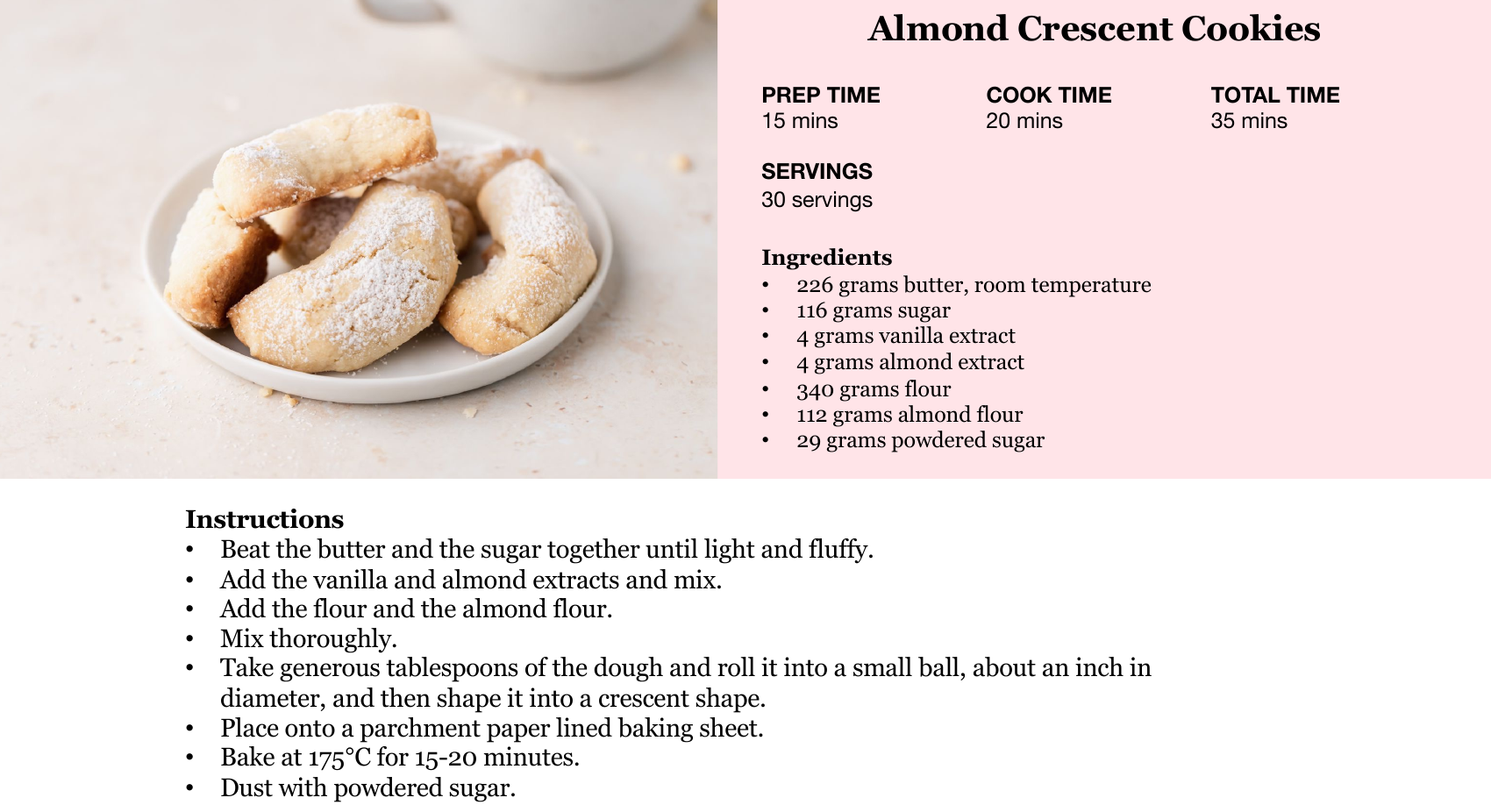}
  \caption{This Figure shows an English recipe for almond crescent cookies, adapted from\\ \url{https://www.simplyrecipes.com/recipes/almond_crescent_cookies/}.}
  \label{fig:recipe}
\end{figure}

We first discuss the main challenges of recipe understanding and related work before introducing \textit{narrative-based understanding} \citep[section \ref{s:narrative-based}, also see][]{vaneecke2023candide}. We then illustrate the approach through a concrete case study on an English recipe for almond crescent cookies, shown in Figure \ref{fig:recipe}. Finally, we evaluate the benefits and the scalability of the approach, and provide more information about resources made available to the community for researchers who wish to experiment with narrative-based understanding.

\subsection{Challenges}
\label{s:challenges}

Recipe understanding is a challenge for robotics because kitchens are rich and dynamically changing environments \citep{bollini2013interpreting}.  From a linguistic perspective, recipes come with their own genre-specific syntax and semantics \citep[see a.o.][]{colleen_claiming_1997,gerhardt_culinary_2013,cani_deconstructing_2022} that challenge traditional NLP solutions, of which we summarize the most important ones here:

\begin{itemize}
\item \textit{How-to instructions:} Recipes use procedural language, such as imperative commands, which leads to reduced performance of off-the-shelf parsers \citep{tellex_robots_2020}.

\item \textit{Zero anaphora:} Recipes are abundant with zero anaphora (e.g. no direct object in the phrase ``mix thoroughly'') because cooking takes place in an actual kitchen that provides the necessary context for filling in the blanks.

\item \textit{Dynamic Environment:} Kitchens are dynamic environments in which entities are changed into ``resultant objects'' -- often without explicit mention of that happening. For instance, the almond cookie recipe (Figure \ref{fig:recipe}) introduces the phrase ``the dough'' for the first time in its fifth instruction without making explicit that it is the resultant object of mixing together various ingredients such as butter, sugar and flour.

\item \textit{Complex Semantics:} Recipes require careful management of time, measurement and ordering. Instructions can be explicit (such as ``340 grams flour'' or ``for 15-20 minutes''), but recipes also often use vague measurements (``generous tablespoons'') and evaluative phrases (``until light and fluffy'') that require a tight integration of language processing and sensorimotor perception.
\end{itemize}

\subsection{Related Work}

Computational recipe understanding and other tasks in \textit{Digital Gastronomy} \citep{zoran_cooking_2019} have always enjoyed academic interest \citep[see e.g. the Computer Cooking Contests;][]{najjar_computer_2017}, but especially in the past few years there has been a surge of attention for the broader field of \textit{food computing} \citep{harper_opanag_2015,min_survey_2019}. This surge is driven on the one hand by the explosion of large-scale online data such as recipes and cooking videos; and on the other hand by the breakthroughs in deep learning for handling such large data \citep[e.g.][]{lecun_deep_2015}. 

Most research therefore focuses on aggregating and cleaning up the data; and on the creation of datasets, benchmarks, representations and classification systems for food-related information \citep[e.g.][]{smith_automatic_2012,kicherer_what_2018,yagcioglu-etal-2018-recipeqa,marin2019learning,popovski_foodbase_2019,jiang-etal-2020-recipe,tian_recipe_2021}. This information is then used for various tasks such as recipe generation \citep[e.g.][]{jabeen_autochef_2020,wang2022learning}, recipe recommendation \citep[e.g.][]{haussmann_foodkg_2019,tian_reciperec_2022}, question-answering systems \citep[e.g.][]{manna_information_2020,khilji_cookingqa_2021}, and so on. Ultimately, such systems aim to provide an appropriate response to a particular input, such as proposing relevant recipes based on the user's preferences.

Even though such work is relevant to the present study, their goals only require a shallow understanding of recipes, while our objective is to parse recipes in such a way that a robot can successfully execute it (or more generally speaking, that the robot can successfully execute instructions). This objective requires adequate systems for \textit{grounded} \citep{harnad_symbol_1990} \textit{natural language understanding} (NLU; \citealp{allen_nlu_1994}, also see \citealp{tellex_robots_2020}, for a survey). 

Despite a longstanding research history going back to the 1970s \citep[e.g.][]{shrdlu,shakey}, a recent benchmark study has shown that grounded language understanding is still a largely unsolved problem \citep{shridhar_alfred_2020}. That is not to say that no progress has been made: thanks to more sophisticated language technologies and the increasing availability of online data, the field has moved away from limited sets of natural language instructions, and has instead set its ambition on mapping open-ended instructions from the web onto everyday manipulation tasks \citep{tenorth_understanding_2010}. 

In the cooking domain, several prototypes and experiments have been reported \citep{sugiura_cooking_2010,beetz_robotic_2011,bollini2013interpreting,bezalelimizrahi_digital_2023}. These studies are usually performed from the perspective of robotics, and mainly examine how existing NLP techniques can be repurposed for the generation of executable robot plans \citep{tenorth_understanding_2010}. Language processing therefore typically involves translating instructions onto syntactic parse trees from which semantics can be inferred; or more recently, applying neural network models for directly mapping sentences onto formal semantic representations \citep{tellex_robots_2020}.

\section{Narrative-Based Understanding}
\label{s:narrative-based}

We propose to treat recipes as a form of narrative, taking inspiration from discourse-analysis studies in linguistics \citep{colleen_claiming_1997} and recent work on the value of narratives for human-centered AI \citep[e.g.][]{szilas:OASIcs:2015:5287,blin_building_2022,steels_foundations_2022,vaneecke2023candide}. Narratologists divide a narrative into three interconnected layers \citep{bal_narratology_1985}, illustrated in Figure \ref{fig:narrative}:

\begin{figure}
\centering
	\includegraphics[width=.6\textwidth]{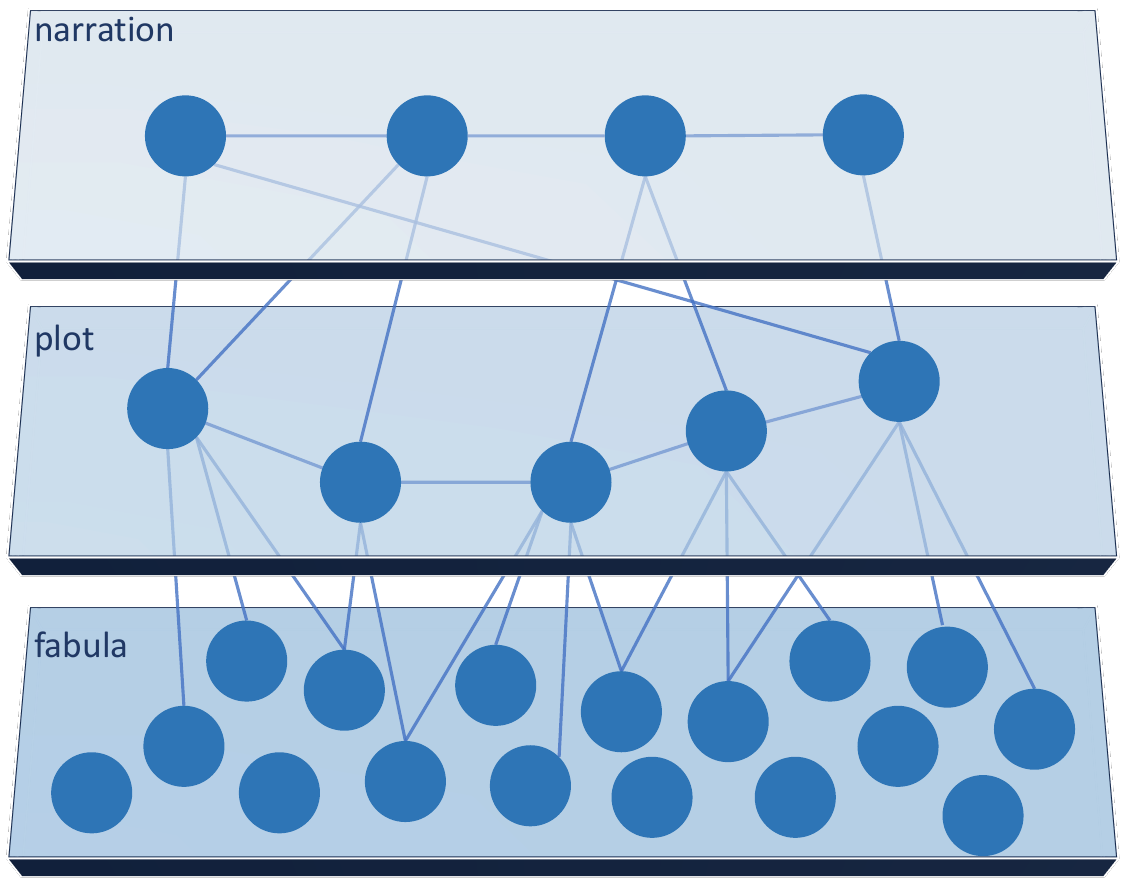}
	\caption{A narrative is a three-layered structure consisting of a fabula, plot, and narration. Narrative-based understanding involves constructing the plot using the narration and the fabula, thereby integrating language processing, memory, mental simulation, perception, and so on.}
	\label{fig:narrative}
\end{figure}

\begin{enumerate}
	
	\item The \textit{fabula} (or \textit{story}) is a collection of facts, events and actions;
	
	\item The \textit{plot} (or \textit{syuzhet}) is a structure that arranges the relevant items of the fabula in a causal network of events that lead to a conclusion \citep[called \textit{narrative closure};][]{carroll_narrative_2007};
	
	\item The \textit{narration} (or \textit{discourse}) is how the narrative is presented.
	
\end{enumerate}

\subsection{Recipes as Narratives}

Let us apply these three layers to recipes-as-narratives. There are two observable layers: the \textit{fabula} and the \textit{narration}. The \textit{fabula} is so vast (i.e. most of its content is irrelevant such as ingredients that won't be used) that a cooking agent can only maintain a partial model, which it obtains through sensorimotor processing and retrieving facts from memory (e.g. which drawer contains the cutlery). While the fabula can be considered as the background against which the narrative should be situated, the \textit{narration} concerns how the narrative is presented, which can be a written recipe, a cooking video, a dialogue, and so on. This layer is typically analyzed using (multimodal) language processing techniques.

At the heart of narrative-based understanding is the \textit{plot}, which is invisible to the cooking agent and which therefore has to be constructed. The plot is a rich content model in which the relevant elements of the fabula are arranged in a causal network of events. By integrating the diverse and often fragmented and ambiguous input from various knowledge sources (such as language processing, vision and pattern recognition, mental simulation, action monitoring, ontologies, knowledge graphs, and so on), the plot provides a coherent and structured path towards the goal of the narrative (in our case study: delicious almond crescent cookies).

In the case of cooking, our main guideline for constructing the causal network of events is the narration. In the simplest case, the narration follows the same order as the plot, but even for recipes there exist many variations \citep{colleen_claiming_1997}. The recipe for almond cookies, for instance, starts with a flash-forward by stating that there will be 30 servings. Other recipes may rely on base recipes (subplots) or include alternative ways to prepare a dish.

Important to note is that the construction of the plot is not the main goal of narrative-based understanding: it rather serves to find \textit{narrative closure} \citep{carroll_narrative_2007}, which is the state in which the plot arrives at a satisfactory conclusion. In the case of recipe understanding, narrative closure is obviously achieved if the desired food is ready to be served.

\subsection{Language as a Form of Action}
\label{s:grammar}

Just like narratives involve the active construction of a plot in order to make sense of reality \citep{bruner1991narrative}, functional theories of linguistics have considered language to be a form of action ever since the influential works of \citet{wittgenstein1953philosophical}, \citet{austin1962how} and \citet{searle_speechacts_1969}. The instructions found in recipes are textbook examples of such \textit{speech acts}: linguistic expressions that invite the addressee to perform a (mental) action.

In robotics and grounded language understanding, those actions take the form of plans that can be simulated or executed by a robot. Traditional approaches typically involve a pipeline going from linguistic expressions to truth-conditional semantic representations \citep{Eckardt_truth_2006,tellex_robots_2020}, which are then mapped onto an executable robot plan. For instance, the phrase ``take the dough'' can be associated with the logical form $\exists x: \{ DOUGH(x) \land TAKEN(x) \}$ (``there exists an \textit{x} that is dough and that is taken''), which \citep[using temporal logics;][]{kress_synthesis_2018} can be specified as becoming True once the robot executes the correct operation and takes the dough. This formal semantic specification is then used for generating an executable robot plan \citep[e.g.][]{beetz_robotic_2011,bollini2013interpreting,sugiura_cooking_2010}.

In our approach, we dispense with an intermediary truth-conditional representation and propose that the meaning of a sentence (or indeed, the meaning of the recipe as a whole) \textit{is} an executable robot plan. For example, the phrase ``take generous tablespoons of the dough and roll it into a small ball'' in the almond crescent cookie recipe directly maps onto an actual operation in which the cooking agent uses a tablespoon as a tool of measurement for making several spheres made of dough.

\subsection{Personal Dynamic Memory}

Narratives are \textit{personal} as they are based on past experiences, individual beliefs and values, and on which perspective is taken \citep{steels_pdm_2020,vaneecke2023candide}. For instance, if at the world cup football a small nation eliminates one of the tournament's favourites, their supporters may praise their team's courage and efficient counter tactics, while the losing side might condemn them for playing a defensive ``anti-football'' game.

Narratives are therefore not constructed out of the blue, but are integrated into a \textit{personal dynamic memory} \citep[PDM;][]{steels_pdm_2020}. A personal dynamic memory consists of persistent knowledge (e.g. an agent's linguistic inventory, its ontology, and so on) and of past experiences and past narratives. The more cooking experience an agent has acquired throughout its lifetime, the easier it will be able to construct a recipe's narrative.

\section{Almond Crescent Cookies}

There is an English proverb that goes \textit{the proof is in the pudding}, which comes from the older saying \textit{the proof of the pudding is in the eating}. This expression used to mean quite literally that you have to try out food to know how tasty it is, and nowadays it can be used to say that you can only know the value or quality of something through direct experience or by obtaining concrete results. The same goes for evaluating the value of our narrative-based approach to recipe understanding.

\begin{figure}
\centering
\includegraphics[width=.9\textwidth]{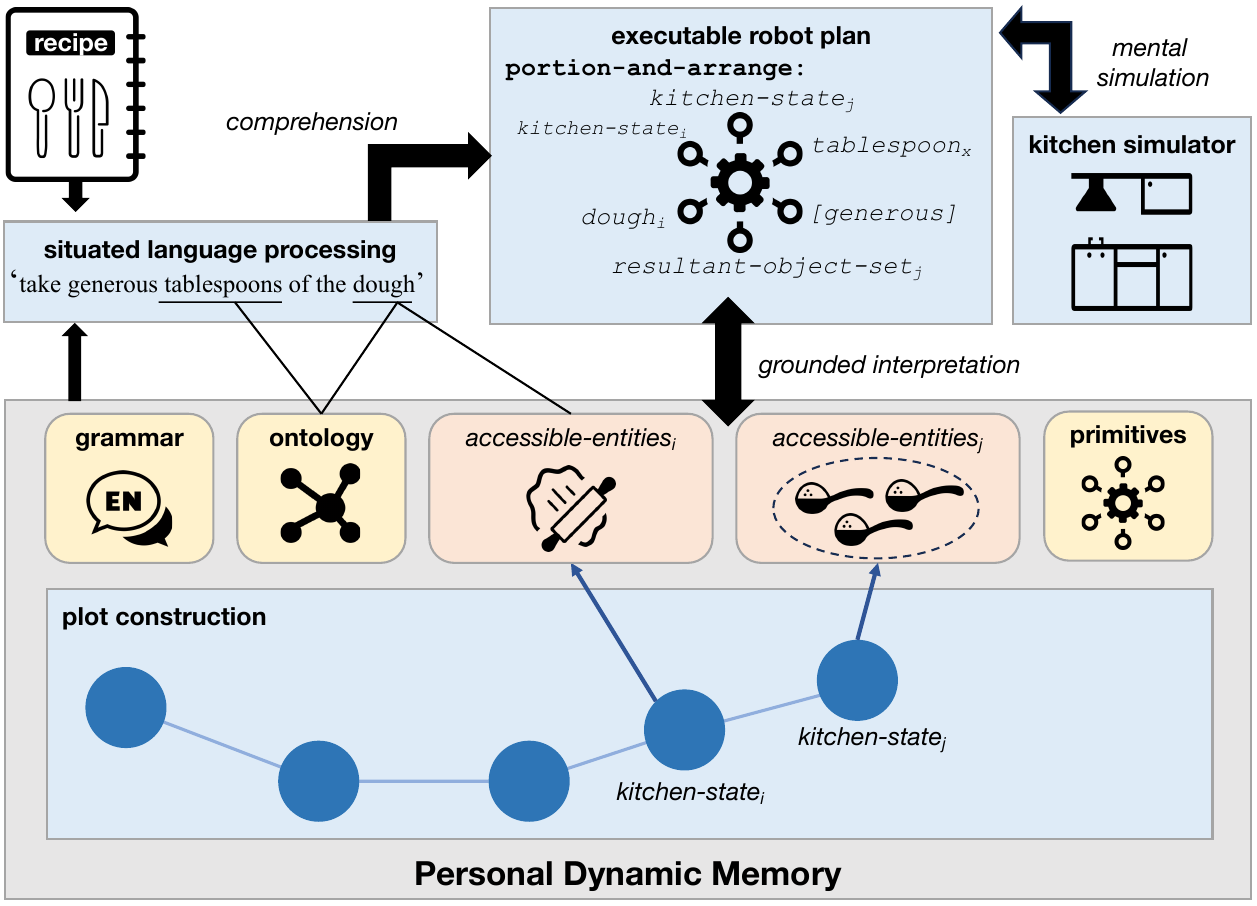}
  \caption{This figure illustrates a single cycle in the construction of the recipe's plot. On the top left: while parsing, the language processor has access to the cooking agent's grammar, ontology, and the entities that are currently under its attention (\textit{accessible-entities$_i$}). Comprehension results in a partial executable robot plan (here the operation \textit{portion-and-arrange}). Through interaction with a kitchen simulator and the agent's personal dynamic memory, a complete plan is generated and executed, leading to a new plot beat (\textit{kitchen-state$_j$}), which includes new accessible entities (tablespoons of dough). The cycle can then repeat itself with the next instruction until the recipe is finished.}
  \label{fig:cycle}
\end{figure}

While the previous section offered a conceptual, implementation-independent overview of narrative-based understanding, we will now proceed with a specific operationalization through a concrete case study on an English recipe for almond crescent cookies. This section will make heavy use of Figure \ref{fig:cycle}, which illustrates one cycle of the back-and-forth between language processing, semantic interpretation, mental simulation, and personal dynamic memories. Interested readers can find more technical details at the web demonstration of our case study at \url{https://ehai.ai.vub.ac.be/demos/recipe-understanding}, and the recipe execution benchmark which we developed for evaluating our approach (see section \ref{s:evaluation}).

\subsection{Situated Language Processing}

In order to handle the genre-specific challenges of recipes, we have chosen a construction grammar approach \citep{fillmore1988mechanisms,goldberg2003constructions,fried2004construction}, which we implemented in the open-source formalism Fluid Construction Grammar \citep[FCG;][]{steels2004constructivist,vantrijp2022fcg,beuls2023fluid}.  The motivation for this approach is threefold. First, construction grammar is a linguistic theory in which \textit{all} linguistic information is represented as mappings between form and meaning (called ``constructions''), which makes it convenient to represent both the idiosyncrasies of recipe language as well as its more abstract syntactic structures in a uniform way. Secondly, semantics can but needn't be directly coupled to syntactic structures, which makes it possible to parse sentences directly into language-independent executable robot plans. Finally, the functional scope of constructions is not limited to the sentence level, which means that constructions can represent discourse-level information as well \citep{fried_discourse-referential_2021}.

The latter feature of construction grammar is of great importance for recipes. As discussed in section \ref{s:challenges}, recipes are abundant with zero anaphora, which are used by recipe authors as a strategy for cohesion building since these zero anaphora refer to entities that are highly salient in the current discourse context \citep{cani_deconstructing_2022}. For instance, in the almond cookies recipe, the direct object is omitted in phrases such as ``mix thoroughly'' and ``place onto a parchment paper lined baking sheet''.

Our solution is to include non-linguistic information in language processing, which is supplied by the personal dynamic memory as shown in Figure \ref{fig:cycle}. The PDM is where the recipe's plot is constructed: at each node, the PDM tracks which entities are currently under the attention of the cooking agent (called \textit{accessible entities}). Accessible entities are like characters that were introduced in prior scenes and that are still present in the current scene.

\begin{figure}
\centering
	\includegraphics[width=.9\textwidth]{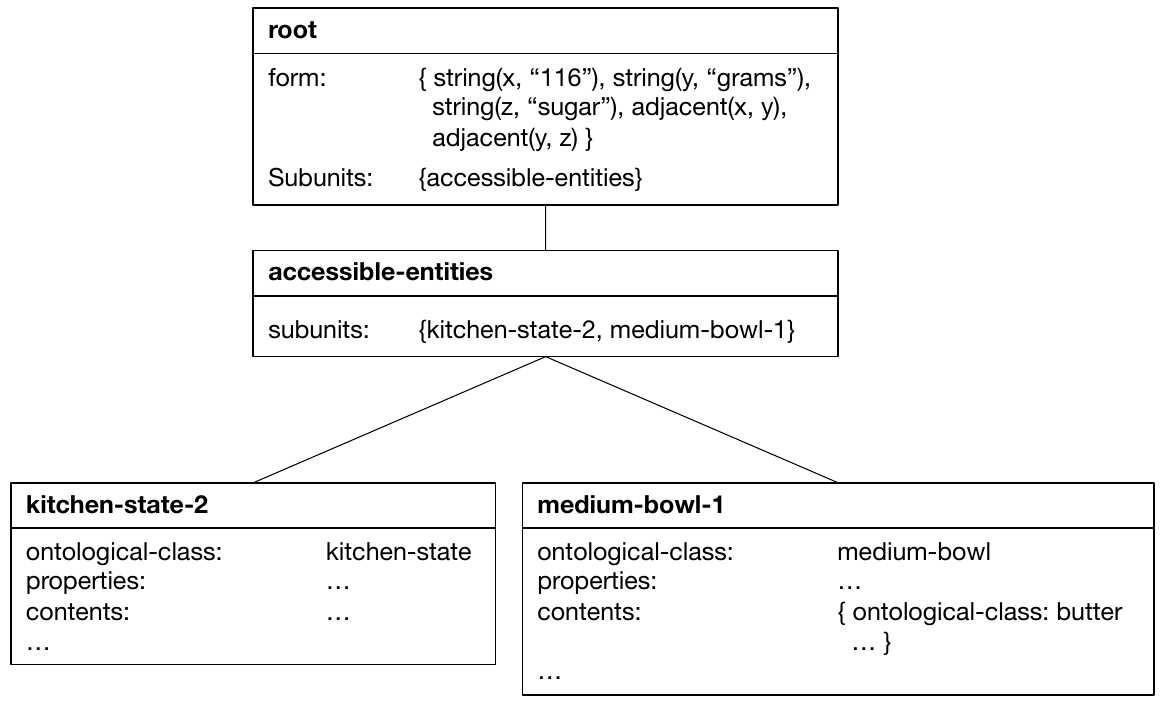}
	\caption{A \textit{transient structure} contains information about both the input sentence and the entities that are accessible from discourse context.}
	\label{fig:transient-structure}
\end{figure}

More concretely, linguistic processing makes use of a kind of blackboard that contains all of the information about the accessible entities and the input phrase or sentence. This blackboard is called \textit{transient structure} because it changes over time as different constructions access and expand its information. Figure \ref{fig:transient-structure} illustrates such a transient structure at the beginning of a parsing task for the phrase ``116 grams sugar''. This transient structure consists of four units (which are simply lists of feature-value pairs). The unit called \textit{root} (on top) contains unhandled information from the input sentence, such as which strings (or tokens) were observed and which strings are adjacent to each other. At this point in the recipe, the cooking agent has already fetched a medium bowl of butter, so there are two accessible entities: the current kitchen state, and the bowl of butter.

In Fluid Construction Grammar (FCG), constructions are formalized as schemas that consist of a conditional pole (right-hand side) and a contributing pole (left-hand side). A construction is allowed to add the information of its contributing pole to the transient structure if the feature-value pairs from its conditional pole can be matched against the information already present in the transient structure \citep{steels2006unify}. Here is an example of a lexical construction for the word ``sugar'':

\begin{center}{\includegraphics[width=.9\textwidth]{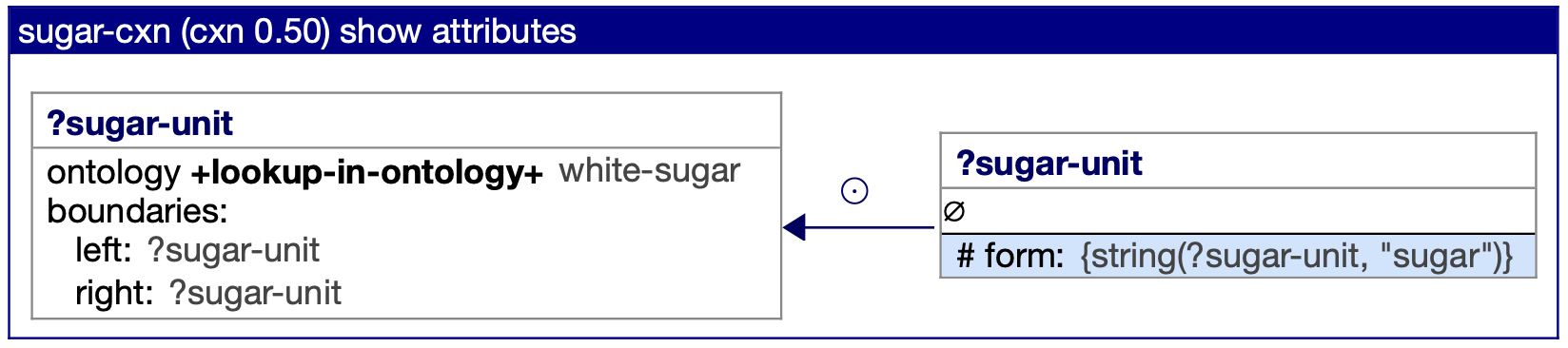}}\end{center}

The application of a construction leads to a new and expanded transient structure, which may in turn trigger the application of other constructions. Parsing is thus operationalized as a search problem for finding the best set of constructions for extracting the meaning of an input sentence \citep{vaneecke2022neural}. 

Constructions represent most information as declarative features, but they can also use procedural attachment \citep{steels_procedural_1979,bundy_procedural_1984}. For instance, in the \textsc{sugar-cxn}, the value of the feature \textit{ontology} is not directly specified: instead, a procedure \texttt{+lookup-in-ontology+} is embedded, which is able to fetch all of the features that are associated with the concept [white-sugar]. Procedural attachment is necessary for dealing with uncertainty. Some examples are:

\begin{enumerate}

	\item Human-centric AI systems must be open-ended learners, hence the ontology may change with every novel experience.
	
	\item Higher-level constructions may involve generalizations, e.g. over phrases with ingredients such as ``two tablespoons of sugar'' or ``120 grams of flour''. Procedural attachment allows an on-the-fly check whether sugar or flour both qualify as ingredients.
	
	\item Constructions that handle quantities must be able to parse such units on the fly. For instance, a construction responsible for portions should recognize both the tokens ``114'' as well as ``two'', so a procedure for checking whether a token can be parsed as a number helps to make such generalizations possible.

\end{enumerate}

For the present case study, we hand-coded a small grammar of 56 constructions that can be inspected in our web demo. Hand-coding is a necessary first step to identify what kind of grammatical structures are necessary for mapping recipes onto executable robot plans and for evaluating the feasibility of a constructional approach (we will address the question of learning in section \ref{s:conclusion}). Our grammar includes constructions for lemmatization, lexical constructions, idiomatic and semi-schematic constructions (e.g. the ``until light and fluffy''- and ``place-X-onto-y''-constructions), and abstract constructions such as the English Resultative \citep{boas_constructional_2003,goldberg_english_2004} for analyzing phrases such as ``beat the butter and sugar together until light and fluffy''.

\subsection{Meaning and Mental Simulation}
\label{s:meaning}

The goal of language processing is not to derive the most accurate syntactic analysis of a sentence: syntactic structures are only built insofar as they help to get to the meaning as efficiently as possible. ``The'' meaning is a bit misleading because language is an inferential coding system \citep{sperber1986relevance} so not all the meaning is in the message. For instance, a phrase such as ``put two eggs in a bowl'' does not specify which eggs and bowl to take, or that you have to crack the eggs open and get rid of the shells.

\begin{figure}
	\begin{center}{\includegraphics[width=0.6\columnwidth]{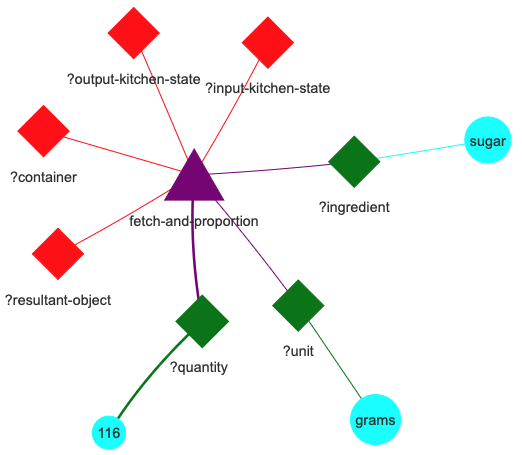}}\end{center}
	\caption{Parsing ``116 grams sugar'' leads to a partial robot plan that the cooking agent needs to complete into an executable one.}
	\label{fig:primitive}
\end{figure}

Parsing therefore only leads to a partial robot plan that the cooking agent needs to complete and expand upon. For instance, the phrase ``116 grams sugar'' maps onto an operation called \textit{fetch-and-proportion}, illustrated in Figure \ref{fig:primitive}. Some of the operation's arguments are already provided by the recipe instructions such as which food product to fetch (\textit{sugar}), shown as cyan circles. However, there are several open slots such as the resultant object, shown as red diamonds. The cooking agent thus needs to find fillers for those slots using different knowledge sources such as its personal dynamic memory and mental simulation. To enable the agent to do so, we used the open-source software tool Incremental Recruitment Language (IRL) for representing, generating and executing robot plans \citep{steels2000emergence,spranger2012open}. 

More specifically, we implemented a new representation language for cooking that includes 40 predefined cooking operations (called ``primitives'') that encode meaning, temporality and dependencies. The IRL-system can then combine these primitives into graphs that represent complete recipe execution plans.  Recurrent graphs can be automatically chunked and stored as composite operations for more efficient plan generation in the future, and users may extend the representation language with additional cooking operations.

As shown in Figure \ref{fig:cycle},  plan execution relies on mental simulation, which is a distinct human capacity that allows us to project ourselves into hypothetical realities \citep{waytz_mental_2015}, typically in the form of narratives \citep{escala_imagine_2004}. In our case study, we therefore included our own (qualitative) kitchen simulator as well as a quantitative simulator \citep{pomarlan_abesim_2021} for simulating cooking operations and their effects (see section \ref{s:benchmark}).

One of the reasons for selecting the IRL-system for plan generation and plan execution is that it allows a \textit{data flow} approach in which the handling of data adapts to which data is available at a given moment, as opposed to explicit control flow in which all operations need to be ordered beforehand. For example, given an initial kitchen state, a food product and the unit of measurement, the operation \textit{fetch-and-proportion} can compute the resultant object and output kitchen state. Suppose however that a human chef has already taken a cup of 116 grams of sugar, then the agent could apply the same operation in a different direction for verifying whether the resultant object corresponds to what is written in the recipe, or for backtracking which actions the human user must have taken.

Data flow is important for handling the dynamic nature of a kitchen environment (where many things can go wrong) and for adapting to the user's needs, who may use different variations of a recipe or who may have different preferences about which actions they like to perform themselves and which to delegate to their AI assistant. Moreover, data flow allows the final robot plan to be greatly optimized. For instance, the cooking agent does not need to wait until an operation such as \textit{boil} is completely finished: the IRL-system will already provide a placeholder for the resultant object (including a timing for when it will be ready) so the cooking agent can already start planning and executing other tasks.

The result of plan generation and execution is a new node in the plot that the agent is constructing in its personal dynamic memory, as shown in Figure \ref{fig:cycle}. This new node includes an update of the kitchen state and which entities are currently accessible. Indeed, the ``meaning'' of each recipe instruction is a small executable robot plan that is causally linked to the next one. The interlinked plans as a whole form a detailed and coherent robot execution plan for the whole recipe, which makes it possible to backtrack to earlier kitchen states whenever necessary.

\section{Evaluation and Self-Assessment}
\label{s:evaluation}

In this section we describe the steps taken towards evaluation as well as first results. Moreover, we discuss how a cooking agent may reason about its own performance.

\subsection{Integrative Narrative Networks}

\begin{figure}
	\centering \includegraphics[width=0.5\textwidth]{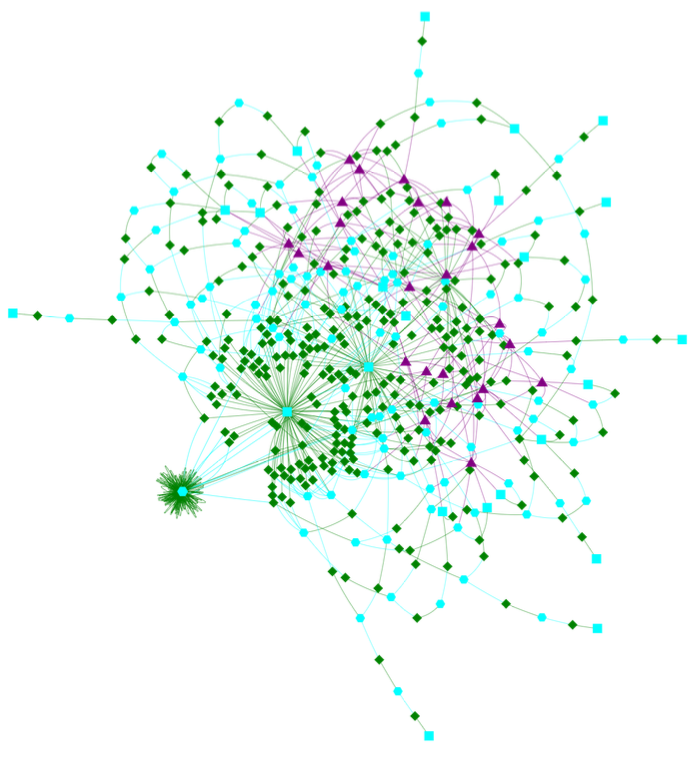}
	\caption{An Integrative Narrative Network for the almond crescent cookies recipe.}
	\label{fig:inn}
\end{figure}

One of the challenges of narrative-based understanding is to monitor how different knowledge sources are integrated with each other, and whether the resulting plot offers a coherent and sensible content model. Crucially, the cooking agent itself should also have a way to monitor its own understanding process. 

We are approaching this challenge using a new data structure called \textit{integrative narrative networks} \citep{baroncini2023semantic}. The key inspiration for such networks comes from the narratological concept of ``narrative questions'', which are the questions that are raised in the audience's mind by a narrative (or that an author wants to be raised). For instance, when a new character is introduced in a movie, this may raise the narrative question ``who is this person?'' Compelling narratives use such questions to connect plot points to each other and to keep the audience engaged, until narrative closure is reached -- the point where all salient questions have been satisfactorily answered \citep{carroll_narrative_2007}. Likewise, we can frame understanding as a process in which an agent poses narrative questions to itself, and then searches for answers for those questions until it reaches narrative closure (or a conclusion).

The network in Figure \ref{fig:primitive} illustrates this idea. This network shows that the activation of the primitive \textit{fetch-and-proportion} raises a number of questions (depicted as diamond-shaped nodes), such as which ingredient to fetch and what the resultant object will be. Some of these questions are immediately answered by parsing ``116 grams sugar'' (the green nodes), while other questions are still open (red). The agent now has to search for answers for those questions. Narrative questions and answers can be introduced by various knowledge sources, and this process is continued until all salient questions have been satisfactorily answered. Figure \ref{fig:inn} shows a complete Integrative Narrative Network, as built by the cooking agent, for the almond crescent cookies recipe.

\begin{figure}
\centering
  \includegraphics[width=.6\textwidth]{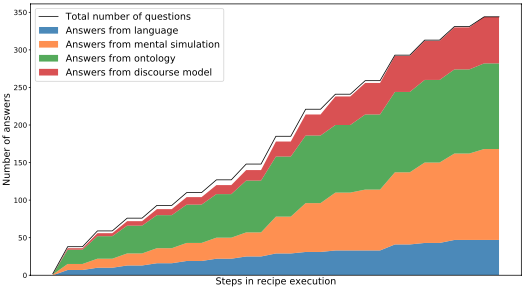}
  \caption{This Figure illustrates the number of narrative questions raised during recipe understanding and how many of them have been answered by which knowledge source.}
  \label{fig:measuring}
\end{figure}

Figure \ref{fig:measuring} shows some results of a first experiment in monitoring and measuring narrative-based understanding, described in more detail in \citet{steels2022experiment}. The black line shows the number of narrative questions that are raised as the cooking agent goes through the recipe. The coloured segments show how many questions were answered by different knowledge sources. From bottom to top, these are: language (blue), mental simulation (orange), ontology (green) and the discourse model/PDM (red).

Our current work focuses on how such integrative narrative networks can be used for allowing agents to monitor and reason about their own understanding process, and to optimize the decisions that they make in the face of uncertainty.

\subsection{Recipe Execution Benchmark}
\label{s:benchmark}

In order to evaluate our work as well as invite the research community to advance the field of grounded language understanding (and computational recipe understanding in particular), we have released a fully documented recipe execution benchmark \citep{nevens2024benchmark}, which consists of the following components:

\begin{itemize}
	\item A representation language for cooking (see section \ref{s:meaning}) that can express complete recipe execution plans. This representation language is independent from syntax or a particular natural language, so knowledge about syntax is not necessary for annotation.
	
	\item A test set of 30 English recipes with gold standard annotations. These recipes have been selected for the specific linguistic and extralinguistic challenges in recipe understanding.
	
	\item A qualitative kitchen simulator that is able to execute the recipe execution plans, and which returns both execution and evaluation results for further inspection.
	
	\item A suite of metrics that allow multiperspective estimates to optimize transferability to real-world utility. These consist of Smatch \citep{cai-knight-2013-smatch}, goal-condition success, Dish Approximation Score, and Recipe Execution Time.
\end{itemize}

\section{Conclusions and Future Work}
\label{s:conclusion}

This paper explored how narrative-based language understanding can be used for processing cooking recipes in a way that allows robots or artificial cooking agents to execute those recipes in a dynamic kitchen environment. Through a case study on an English recipe for almond crescent cookies, we have shown how the rich content models built during narrative-based understanding can be exploited for tackling the specific challenges of recipe language, such as resolving zero anaphora by keeping track of which entities are currently under the cooking agent's attention. We have thereby shown how language processing can be embedded in a system for representing, generating and executing robot plans, coupled to a kitchen simulator. Moreover, we have proposed how narratives may offer a new framework for monitoring and measuring understanding.

Even though we have illustrated our approach through a concrete implementation and accompanying web demonstration, we hope to have convinced the reader that the framework of narrative-based understanding can be operationalized in a multitude of ways. To this end we have published a recipe execution benchmark for comparing different solutions. One key component of this benchmark is a new representation language for cooking, which allows recipes to be annotated in a syntax- and natural-language-independent fashion.

Our current and future work focuses on automatically learning computational construction grammars for recipe understanding in order to scale our approach. For the reasons detailed in section \ref{s:grammar}, we believe that construction grammar shows great promise to deal with the specific challenges of grounded language understanding, and computational recipe understanding in particular. Important breakthroughs in the automated learning of such grammars have recently been achieved in the domain of Visual-Question Answering \citep{nevens2022language,doumen2024modelling,beuls2024humans}, which requires a mapping from questions to visual queries in similar ways as recipe instructions map onto executable robot plans.

By mimicking the sense-making process of humans, narrative-based language understanding can become a key component in the development of human-centric AI systems. Such systems have many potential benefits for society, as they may interact with humans in more intuitive and meaningful ways.

\section*{Acknowledgements}

The research reported on in this chapter was funded by the European Union’s Horizon 2020 research and innovation programme under grant agreement no. 951846 (MUHAI).

\bibliographystyle{apalike}
\bibliography{refs-almond,../../../../ehaidocs/Bibliography/ehai_bibliography}

\end{document}